\newcommand{\fref}[1]{Fig.~\ref{#1}}
\newcommand{\tref}[1]{Table~\ref{#1}}
\newcommand{\sref}[1]{Section~\ref{#1}}

\documentclass{article}
\usepackage{spconf,amsmath,graphicx}
\usepackage{multirow}
\usepackage{algorithm}
\usepackage[noend]{algpseudocode}
\usepackage{xcolor}
\usepackage{enumitem}
\usepackage[misc]{ifsym}

\title{GNP Attack: Transferable Adversarial Examples via Gradient Norm Penalty}
\name{Tao Wu$^1$, Tie Luo$^1$\textsuperscript{\Letter}, Donald C. Wunsch$^2$\vspace{-3mm}}
\address{$^1$Department of Computer Science, $^2$Department of Electrical and Computer Engineering\\
Missouri University of Science and Technology\\
\{wuta, tluo, dwunsch\}@mst.edu\vspace{-3mm}}
\begin{document}
%
\maketitle
\begin{abstract}
Adversarial examples (AE) with good transferability enable practical black-box attacks on diverse target models, where insider knowledge about the target models is not required. Previous methods often generate AE with no or very limited transferability; that is, they easily overfit to the particular architecture and feature representation of the \textit{source, white-box} model and the generated AE barely work for \textit{target, black-box} models. In this paper, we propose a novel approach to enhance AE transferability using Gradient Norm Penalty (GNP). It drives the loss function optimization procedure to converge to a flat region of local optima in the loss landscape. By attacking 11 state-of-the-art (SOTA) deep learning models and 6 advanced defense methods, we empirically show that GNP is very effective in generating AE with high transferability. We also demonstrate that it is very flexible in that it can be easily integrated with other gradient based methods for stronger transfer-based attacks.
\end{abstract}
\section{Introduction}
\label{sec:intro}
Deep Neural Networks (DNNs) are the workhorse of a broad variety of computer vision tasks but are vulnerable to adversarial examples (AE), which are data samples (typically images) that are perturbed by human-imperceptible noises yet result in odd misclassifications.
This lack of adversarial robustness curtails and often even prevents deep learning models from being deployed in security or safety critical domains such as healthcare, neuroscience, finance, and self-driving cars, to name a few. 

Adversarial examples are commonly studied under two settings, white-box and black-box attacks. In the white-box setting, adversaries have full knowledge of victim models, including model structures, parameters and weights, and loss functions used to train
the models. Therefore, they can directly obtain the gradients of the victim models and seek adversarial examples by misleading the loss function toward incorrect predictions. White-box attacks are important for evaluating and developing robust models and serve as the backend method for many black-box attacks, but is limited in use due to its requirement of having to know the internal details of target models. In the black-box setting, adversaries do not need specific knowledge about victim models other than their external properties (type of input and output). Two types of approaches, query-based and transfer-based, are commonly studied for black-box attacks. The query-based approach attempts to estimate the gradients of a victim model by querying it with a large number of input samples and inspecting the outputs. Due to the large number of queries, it can be easily detected and defended. The transfer-based approach uses {\em surrogate} models to generate {\em transferable} AE which can attack a range of models instead of a single victim model. Hence it is a more attractive approach to black-box attacks.

This paper takes the second approach and focuses on designing a new and effective method to improve the \emph{transferability} of AE. Several directions for boosting adversarial transferability have appeared. Dong et al. \cite{dong2018boosting} proposed momentum based methods. Attention-guided transfer attack (ATA) \cite{wu2020boosting} uses attention maps to identify common features for attacking. Diverse Input Method (DIM) \cite{xie2019improving} calculates the average gradients of augmented images. \cite{liu2017delving} generates transferable AE using an ensemble of multiple models. 

Despite the efforts of previous works, there still exists a large gap of attack success rate between the transfer-based setting and the ideal white-box setting. 
In this paper, we propose a novel method to boost adversarial transferability from an {\em optimization} perspective. Inspired by the concept of ``flat minima'' in the optimization theory \cite{foret2021sharpnessaware} 
which improves the generalization of DNNs, we seek to generate AE that lie in flat regions where the input gradient norm is small, so as to ``generalize'' to other victim models that AE are {\em not} generated on. In a nutshell, this work makes the following contributions:

\begin{itemize}[nosep]
    \item We propose a transfer-based black-box attack from a new perspective that seeks AE in a flat region of loss landscape by penalizing the input gradient norm.
    \item We show that our method, input gradient norm penalty (GNP), can significantly boost the adversarial transferability for a wide range of deep networks.
    \item We demonstrate that GNP can be easily integrated with existing transfer-based attacks to produce even better performance, indicating a highly desirable flexibility.
\end{itemize}

\begin{figure}[t]
    \centering
    \includegraphics[width=\linewidth]{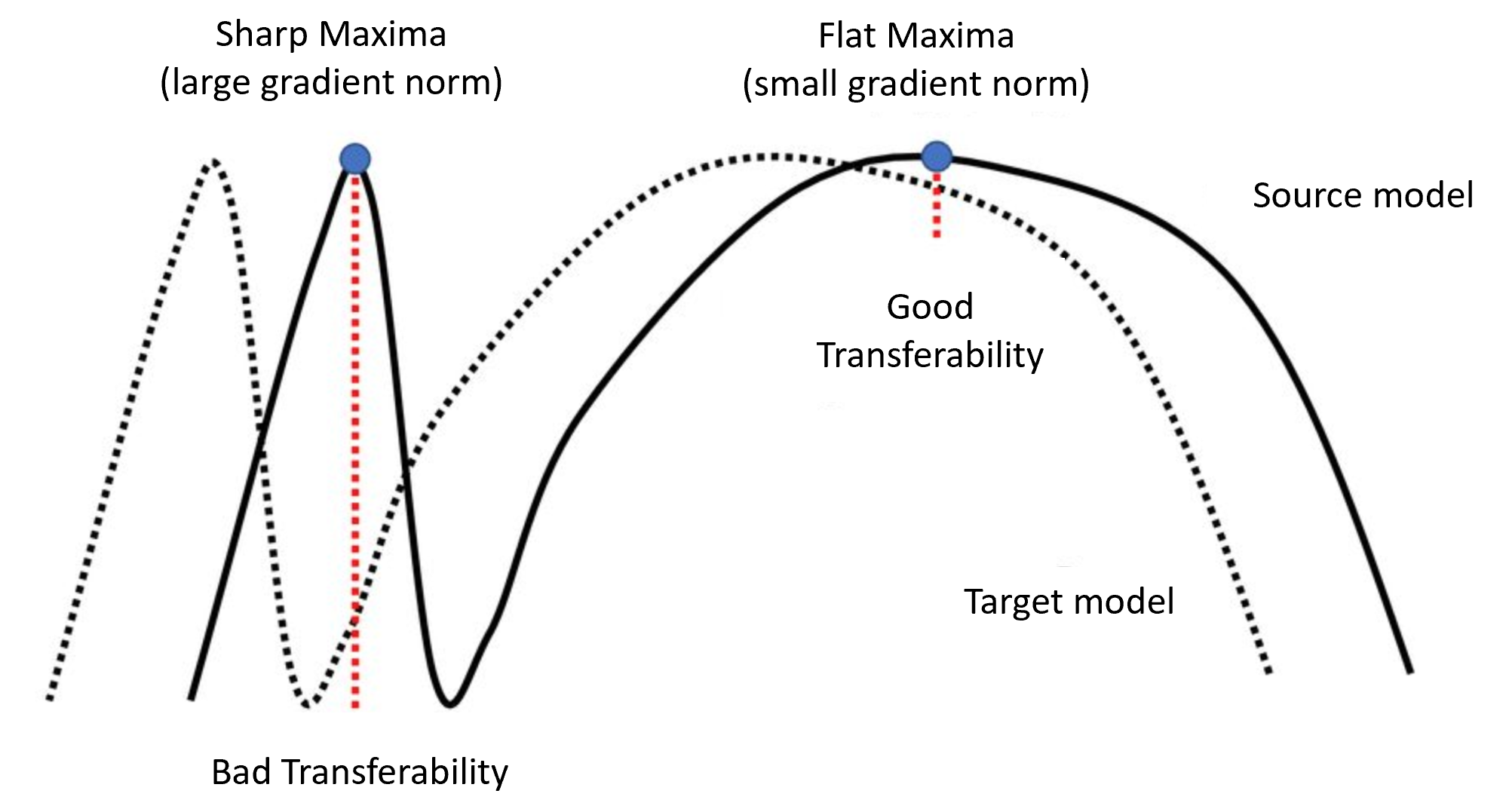}\vspace{-3mm}
    \caption{The loss function landscape: sharpness vs. flatness, which leads to different levels of transferability.}
    \label{fig:diagram}
    \vspace{-3mm}
\end{figure}

\section{Method}\label{sec:method}
\vspace{-2mm}

Given a classification model $f(x): x \in \mathcal{X} \rightarrow y \in \mathcal{Y}$ that outputs a label $y$ as the prediction for an input $x$, we aim to craft an adversarial example $x^*$ which is visually indistinguishable from $x$ but will be misclassified by the classifier, i.e., $f\left(x^*\right) \neq y$. The generation of AE can be formulated as the following optimization problem:
\vspace{-1mm}
\begin{align} \label{eq:1}
    \arg \max _{x^*} \ell\left(x^*, y\right), \;\; \text {s.t. }\left\|x^*-x\right\|_p \leq \epsilon,
\end{align}
where the loss function $\ell(\cdot, \cdot)$ is often the cross-entropy loss, and the $\l_p$-norm measures the discrepancy between $x$ and $x^*$. In this work, we use $p=\infty$ which is commonly  adopted in the literature. Optimizing Eq.~\eqref{eq:1} needs to calculate the gradient of the loss function, but this is not feasible in the black-box setting. Therefore, we aim to create transferable AE on a source model yet can attack many other target models.

We develop a new method to boost adversarial transferability from a perspective inspired by ``flat optima'' in optimization theory. See Fig.~\ref{fig:diagram}. If an AE is located at a sharp local maximum, it will be sensitive to the difference of decision boundaries between the source model and target models. In contrast, if it is located at a flat maximum region, it is much more likely to result in a similar high loss on other models (which is desired).

Thus, we seek to generate AE in flat regions. To this end, we introduce a {\em gradient norm penalty} (GNP) term into the loss function, which penalizes the gradient norm of the loss function with respect to input. The reason is that flat regions are characterized by small gradient norms, hence penalizing the gradient norm will encourage the optimizer to find an AE that lies in a flat region. We thus enhance the adversarial transferability since a minor shift of decision boundary will not significantly change the loss value (prior work has shown that different networks often share similar decision boundaries).

\begin{algorithm}[t]
\caption{I-FGSM+GNP}
\label{alg}
\begin{flushleft}
    \textbf{Input:} A clean sample $x$ with ground-truth label $y$; source model $f(\cdot)$ with loss function $\ell(\cdot)$;\\
    \textbf{Input:} Perturbation size $\epsilon$; maximum iterations $T$; step length $r$; regularization coefficient $\beta$\\
    \textbf{Output:} A transferable AE $x^{adv}$
\end{flushleft}\vspace{-3mm}
\begin{algorithmic}[1]
    \State $\alpha=\epsilon/T$; $g_0=0$; $x_{0}^{adv}=x$
    \For{$t=0$ to $T-1$}
        \State $g_1=\nabla_{x} \ell(x)$
        \State Compute  $r \frac{\nabla_{x} \ell(x)}{||\nabla_{x} \ell(x)||}$
        \State $g_2=\nabla_{x} \ell\left(x+r \frac{\nabla_{x} \ell(x)}{||\nabla_{x} \ell(x)||}\right)$ 
        \State $g_t=(1+\beta)g_1-\beta g_2$
        \State Update $x_{t+1}^{adv} = x_t^{adv} + \alpha \cdot \text{sign}(g_{t})$
    \EndFor
    \State \textbf{return} $x^{adv}=x_{T}^{adv}$
\end{algorithmic}

\end{algorithm}

\vspace{-3mm}
\subsection{Baseline Attacks}

GNP is a very flexible method in that it can be easily incorporated into any existing gradient based method to boost its strength. We consider the following existing, gradient based attacks to demonstrate the effect of GNP. 
Later in \sref{sec:experiments}, we will also show how GNP works effectively on state-of-the-art transfer-based attacks as well.

\textbf{Fast Gradient Sign Method (FGSM).} FGSM \cite{goodfellow2015FGSM} is the first gradient-based attack which crafts an AE $x^{adv}$ by attempting to maximize the loss function $J(x^{adv}, y; \theta)$ with a one-step update:
\begin{equation}
    x^{adv}=x+\epsilon \cdot \text{sign}(\nabla_x \ell(x, y; \theta)),
\end{equation}
where $\nabla_x J(x, y; \theta)$ is the gradient of loss function with respect to $x$, and $\text{sign}(\cdot)$ denotes the sign function.

\textbf{Iterative Fast Gradient Sign Method (I-FGSM).} I-FGSM extends FGSM to an iterative version:
\begin{align}
\label{eq:ifgsm}
    x_{t+1}^{adv} &= x_t^{adv} + \alpha \cdot \text{sign}(\nabla_{x_t^{adv}} \ell(x_t^{adv}, y; \theta)), \\
    x_0^{adv} &= x, \nonumber
\end{align}
where $\alpha=\epsilon / T$ is a small step size and $T$ is the number of iterations.

\begin{table*}[ht!]
\begin{center}
\resizebox{0.999\linewidth}{!}{
\begin{tabular}{|l|c|cccccccccccc|}
\hline
Method  & $\epsilon$ & ResNet50* & VGG19 & ResNet152$\,$ & Inc v3  & DenseNet & MobileNet & SENet  & ResNeXt   & \ WRN\  & PNASNet   & MNASNet  & Average  \\
\hline\hline
\multirow{4}{*}{I-FGSM} & 16/255 & 100.00\% & 61.50\% & 52.82\% & 30.86\% & 57.36\% & 58.92\% & 38.12\% & 48.88\% & 48.92\% & 28.92\% & 57.20\% & 48.35\% \\
 & 8/255  & 100.00\% & 38.90\% & 29.36\% & 15.36\% & 34.86\% & 37.66\% & 17.76\% & 26.30\% & 26.26\% & 13.04\% & 35.08\% & 27.46\% \\
 & 4/255  & 100.00\% & 18.86\% & 11.28\% & 6.66\%  & 15.44\% & 18.36\% & 5.72\%  & 9.58\%  & 9.98\%  & 4.14\%  & 17.02\% & 11.70\% \\
\hline
\multirow{4}{*}{I-FGSM+GNP} & 16/255 & 100.00\% & \textbf{75.96\%} & \textbf{68.89\%} & \textbf{48.23\%} & \textbf{73.68\%} & \textbf{74.05\%} & \textbf{55.46\%} & \textbf{62.36\%} & \textbf{70.60\%} & \textbf{45.06\%} & \textbf{76.98\%} &  \textbf{67.12\%} \\
& 8/255  & 99.96\% & \textbf{68.56\%} & \textbf{60.65\%} & \textbf{38.58\%} & \textbf{62.05\%} & \textbf{63.23\%} &  \textbf{43.69\%} & \textbf{50.36\%} & \textbf{59.32\%} & \textbf{33.62\%} & \textbf{60.28\%} &  \textbf{53.97\%} \\
& 4/255  & 99.98\% & \textbf{25.96\%} & \textbf{22.35\%} & \textbf{15.86\%} & \textbf{26.89\%} & \textbf{28.66\%} & \textbf{15.62\%} & \textbf{21.93\%} & \textbf{23.06\%} & \textbf{13.69\%} & \textbf{30.21\%} &  \textbf{22.38\%} \\
\hline\hline
\multirow{4}{*}{MI-FGSM} & 16/255 & 100.00\% & 73.01\% & 67.62\% & {47.51\%} & {73.16\%} & {72.42\%} & {54.53\%} & {61.78\%} & {60.96\%} & {44.10\%} & {71.46\%} & {62.75\%} \\
& 8/255  & 100.00\% & {52.50\%} & {41.52\%} & {25.56\%} & {47.25\%} & {48.96\%} & {28.06\%} & {35.81\%} & {37.56\%} & {20.41\%} & {47.62\%} & {38.53\%} \\
& 4/255  & 99.94\% & {25.74\%} & {16.68\%} & {9.95\%} & {22.54\%} & {24.89\%} & {9.56\%} & {14.20\%} & {15.38\%} & {7.23\%} & {23.27\%} & {16.94\%} \\
\hline
\multirow{4}{*}{MI-FGSM+GNP} & 16/255 & 100\% & \textbf{89.65\%} & \textbf{83.69\%} & \textbf{65.86\%} & \textbf{87.96\%} & \textbf{90.06\%} & \textbf{69.74\%} & \textbf{79.12\%} & \textbf{77.36\%} & \textbf{58.60\%} & \textbf{88.25\%} & \textbf{79.04\%} \\
& 8/255  & 99.91\% & \textbf{65.28\%} & \textbf{55.63\%} & \textbf{39.69\%} & \textbf{61.42\%} & \textbf{63.26\%} & \textbf{42.03\%} & \textbf{48.65\%} & \textbf{51.07\%} & \textbf{35.03\%} & \textbf{58.93\%} & \textbf{52.20\%} \\
& 4/255  & 100.00\% & \textbf{39.62\%} & \textbf{33.25\%} & \textbf{15.62\%} & \textbf{37.96\%} & \textbf{40.04\%} & \textbf{20.35\%} & \textbf{30.27\%} & \textbf{30.05\%} & \textbf{15.23\%} & \textbf{37.92\%} & \textbf{30.03\%} \\
\hline
\end{tabular}
}
 \caption{Attack success rates when GNP is integrated with baselines to attack 11 target models ('*' denotes white-box attack).}
\label{tab:1}
\vspace{-5mm}
\end{center}
\end{table*}

\textbf{Momentum Iterative Fast Gradient Sign Method (MI-FGSM).} MI-FGSM \cite{dong2018boosting} integrates a momentum term into I-FGSM and improves transferability by a large margin:  
\begin{align}
    g_{t+1} &= \mu \cdot g_t + \frac{\nabla_{x_t^{adv}}J(x_t^{adv}, y; \theta)}{\|\nabla_{x_t^{adv}}J(x_t^{adv}, y; \theta)\|_1}, \label{eq:momentum}\\
    x_{t+1}^{adv} &= x_t^{adv} + \alpha \cdot \text{sign}(g_{t+1}), \nonumber
\end{align}
where $g_0 = 0$ and $\mu$ is a decay factor. 

\subsection{GNP Attack}



As explained in \sref{sec:method}, we aim to guide the loss function optimization process to move into a flat local optimal region. 
To this end, we introduce GNP to penalize large gradient norm, as
\begin{equation}\label{eq:L}
L({x, y}) = \ell(x, y)-\lambda\left\|\nabla_x \ell(x, y)\right\|_2
\end{equation}
where $\ell(\cdot)$ is the original loss function of the source model, and the regularization term is our GNP, which encourages small gradient norm when finding local maxima.

For gradient based attacks (e.g., FGSM, I-FGSM, MI-FGSM, etc.), we need to calculate the gradient
of the new loss \eqref{eq:L}. To simplify notation, we omit $y$ in the loss function since we are calculating gradient with respect to $x$. Using the chain rule, we have\vspace{-2mm}
\begin{equation}\label{eq:grad}
\nabla_{{x}} L({x})=\nabla_{{x}} \ell_{}({x})-\lambda \nabla_{{x}}^2 \ell_{}({x}) \frac{\nabla_{{x}} \ell_{}({x})}{\left\|\nabla_{{x}} \ell_{}({x})\right\|}
\end{equation}
This equation involves the calculation of Hessian matrix $H = \nabla_{{x}}^2 \ell_{}({x})$. This is often infeasible because of the curse of dimensionality (such a Hessian matrix in DNNs tends to be too large due to the often large input dimension). Therefore, we take the first-order Taylor expansion together with the finite difference method (FDM) to approximate the following gradient:
\begin{equation}
\nabla_{{x}} L_{}({x}+r\Delta {x})\approx\nabla_{{x}} \ell_{}({x})+{H} r\Delta {x}
\end{equation}
where $\Delta {x}=\frac{\nabla_x \ell(x)}{\left\|\nabla_{x} \ell(x)\right\|}$, and $r$ is the step length to control the neighborhood size. Thus we obtain the regularization term of \eqref{eq:grad} as: 
\begin{equation}\label{eq:regu}
{H} \frac{\nabla_{{x}} \ell_{}({x})}{\left\|\nabla_{{x}} \ell_{}({x})\right\|} \approx \frac{\nabla_{{x}} \ell\left({x}+r \frac{\nabla_{{x}} \ell_{}({x})}{\left\|\nabla_{{x}} \ell_{}({x})\right\|}\right)-\nabla_{{x}} \ell({x})}{r}
\end{equation}

Inserting \eqref{eq:regu} back into \eqref{eq:grad}, we obtain the gradient of the regularized loss function as:
\begin{equation}
\nabla_{{x}} L({x})=(1+\beta) \nabla_{{x}} \ell_{}({x}) -\beta \nabla_{{x}} \ell_{}\left({x}+r \frac{\nabla_{{x}} \ell_{}({x})}{\left\|\nabla_{{x}} \ell_{}({x})\right\|}\right)
\end{equation}
where $\beta=\frac{\lambda}{r}$ is the regularization coefficient. We summarize the algorithm of how GNP is integrated into I-FGSM in Algorithm \ref{alg}, but I-FGSM can be replaced by any gradient based attack.

\begin{table*}[t!]

\begin{center}
\resizebox{0.999\linewidth}{!}{
\begin{tabular}{|l|c|cccccccccccc|}
\hline
Method  & $\epsilon$ & ResNet50* & VGG19 & ResNet152$\,$ & Inc v3  & DenseNet & MobileNet v2 & SENet  & ResNeXt   & \ WRN\  & PNASNet   & MNASNet  & Average  \\
\hline\hline
\multirow{4}{*}{DIM} & 16/255 &  100.00\% &  93.70\% &  93.62\% &  72.96\% &  94.32\% &  91.68\% &  79.41\% &  91.65\% &  91.17\% &  76.34\% &  89.07\% &  87.47\% \\
 & 8/255  &  100.00\% &  74.01\% &  71.32\% &  40.58\% &  74.65\% &  71.63\% &  44.32\% &  63.38\% &  64.32\% &  40.29\% &  67.27\% &  61.28\% \\
 & 4/255  &  100.00 \% &  39.21\% &  31.65\% &  15.93\% &  38.35\% &  36.74\% &  15.42\% &  25.53\% &  28.68\% &  12.40\% &  33.56\% &  27.76\%  \\
\hline
\multirow{4}{*}{DIM+GNP} & 16/255 &   100.00\% &  \textbf{96.49\%} &  \textbf{97.38\%} &  \textbf{76.89\%} &  \textbf{97.86\%} &  \textbf{95.73\%} &  \textbf{84.56\%} &  \textbf{95.38\%} &  \textbf{96.04\%} &  \textbf{81.69\%} &  \textbf{93.51\%} &  \textbf{91.95\%} \\
& 8/255  &  100.00\% &  \textbf{85.63\%} &  \textbf{84.21\%} &  \textbf{49.65\%} &  \textbf{85.32\%} &  \textbf{80.59\%} &  \textbf{56.24\%} &  \textbf{72.39\%} &  \textbf{75.52\%} &  \textbf{51.68\%} &  \textbf{78.16\%} &  \textbf{72.24\%} \\
& 4/255  &  100.00\% &  \textbf{51.36\%} &  \textbf{45.69\%} &  \textbf{27.96\%} & \textbf{51.39\%} &  \textbf{49.29\%} &  \textbf{28.13\%} &  \textbf{40.08\%} &  \textbf{39.64\%} &  \textbf{25.97\%} &  \textbf{45.23\%} &  \textbf{40.96\%}  \\
\hline
\multirow{4}{*}{TIM} & 16/255 &  100.00\% & 79.90\% &  76.28\% &  54.41\% &  85.42\% &  77.68\% &  55.02\% &  74.15\% &  73.86\% &  62.07\% &  74.38\% &  73.34\% \\
& 8/255  &  100.00\% &  54.91\% &  44.76\% &  28.29\% &  58.17\% &  51.02\% &  24.16\% &  41.70\% &  46.08\% &  29.05\% &  48.92\% &  41.71\% \\
& 4/255 &  99.92\% &  24.31\% &  17.23\% &  12.67\% &  28.42\% &  23.24\% &  6.56\% &  15.03\% &  18.25\% &  9.94\% &  22.76\% &  18.95\% \\
\hline
\multirow{4}{*}{TIM+GNP} & 16/255 &  100.00\% &\textbf{  93.61\%} &\textbf{  90.39\%} &\textbf{  68.43\%} &\textbf{  96.89\%} &\textbf{  91.23\%} &\textbf{  69.01\%} &\textbf{  87.32\%} &\textbf{  84.69\%} &\textbf{  76.25\%} &\textbf{  85.39\%} &\textbf{  84.30\%} \\
& 8/255  &100.00\% &\textbf{  70.03\%} &\textbf{  61.29\%} &\textbf{  45.12\%} &\textbf{  71.35\%} &\textbf{  66.23\%} &\textbf{  41.03\%} &\textbf{  55.46\%} &\textbf{  60.12\%} &\textbf{  46.20\%} &\textbf{ 62.97\%} &\textbf{  57.99\%} \\
& 4/255  & 100.00\% &\textbf{  35.96\%} &\textbf{  35.03\%} &\textbf{  25.16\%} &\textbf{  43.17\%} &\textbf{  36.95\%} &\textbf{  20.36\%} &\textbf{  30.31\%} &\textbf{  32.01\%} &\textbf{  23.68\%} &\textbf{  39.05\%} &\textbf{  32.27\%} \\
\hline
\end{tabular}
}
 \caption{ASR when GNP is integrated with transfer-based attacks to attack 11 target models ('*' denotes white-box attack).}
\label{tab:2}
\end{center}
\end{table*}

\vspace{-5mm}
\section{Experiments} \label{sec:experiments}

\subsection{Experiment Setup} 

{\bf \noindent Dataset and models.} We randomly sample 5,000 test images that can be correctly classified by all the models, from the ImageNet \cite{russakovsky2015imagenet} validation set. We consider 11 SOTA DNN-based image classifiers: ResNet50 \cite{He2016}, VGG-19 \cite{Simonyan2015}, ResNet-152 \cite{He2016}, Inc v3 \cite{Szegedy2016}, DenseNet \cite{Huang2017densely}, MobileNet v2 \cite{Sandler2018mobilenetv2}, SENet \cite{Hu2018}, ResNeXt \cite{Xie2017aggregated}, WRN \cite{Zagoruyko2016}, PNASNet \cite{Liu2018}, and MNASNet \cite{Tan2019mnasnet}. Following the work in~\cite{li2020yet}, we choose ResNet50 as the source model and the remining 10 models as target models.

{\bf \noindent Implementation Details.} In experiments, the pixel values of all images are scaled to [0, 1]. The adversarial perturbation is restricted by 3 scales $\epsilon=4/255,8/255,16/255$. The step length is set as $r=0.01$ and regularization coefficient $\beta=0.8$, we run 100 iterations for all attacks and evaluate model misclassification as attack success rate. 

\begin{table*}[ht!]
\small
\begin{center}
\scalebox{0.999}{
\begin{tabular}{|c|c|cccccc|}
\hline
Source model & Attack & Inc-v3$_{ens3}$ & Inc-v3$_{ens4}$ & IncRes-v2$_{ens1}$ & JPEG & R\&P & NRP \\
\hline\hline
\multirow{2}{*}{ResNet50} 
 & DIM+TIM &  52.13\% & 48.79\% & 38.96\% & 54.85\% & 49.75\% & 39.44\% \\
 & DIM+TIM+GNP & \textbf{65.69\%}  & \textbf{63.16\%}  & \textbf{52.89\%}  &  \textbf{66.31\%} & \textbf{62.04\%} & \textbf{52.81\%} \\
\hline
\end{tabular}
}
\caption{Attacking advanced defense models which underwent adversarial training.}
\label{tab:3}
\vspace{-5mm}
\end{center}
\end{table*}

\vspace{-2mm}
\subsection{Experimental Results}

\subsubsection{Integration with baseline attacks}
We first evaluate the performance of GNP by integrating it with baseline attacks including I-FGSM and MI-FGSM. The results are shown in \tref{tab:1}. We use a pre-trained ResNet50 as the source model and evaluate the attack success rate (ASR) of the generated AE on a variety of target models under different scales of perturbation $\epsilon$. GNP achieves significant and consistent improvement in all the cases. For instance, taking the average ASR of all the 10 target models under perturbation $\epsilon = 8/255$, GNP outperforms I-FGSM and MI-FGSM by 26.51\% and 13.67\%, respectively. In addition, the improvements of the attack success rates on a single model can be achieved by a large margin of 33.06\%.

\subsubsection{Integration with existing transfer-based attacks}
Here we also evaluate the effectiveness of GNP when incorporated into other transfer-based attacks such as DIM \cite{dong2019evading} and TIM \cite{xie2019improving}. The results are given in \tref{tab:2} and show that DIM+GNP and TIM+GNP are clear winners over DIM and TIM alone, respectively. Specifically, DIM+GNP achieves an average success rate of 91.95\% under $\epsilon = 16/255$ for the 10 target models, and TIM+GNP outperform TIM by a large margin of 16.28\% under $\epsilon = 8/255$. We note that we only present the integration of GNP with two typical methods here, but our method also apply to other more powerful gradient-based attack methods.

\subsubsection{Attacking ``secured'' models}
For a more thorough evaluation, we also investigate how GNP will perform when attacking DNN models that have been {\em adversarially trained} (and hence are much harder to attack). We choose three such advanced defense methods to attack, namely, JPEG \cite{guo2018countering}, R\&P \cite{xie2018mitigating} and NRP \cite{naseer2020NRP}. In addition, we choose another three {\em ensemble} adversarially trained (AT) models, which are even harder than regular AT models, and attack them: Inc-v3$_{ens3}$, Inc-v3$_{ens4}$ and IncRes-v2$_{ens1}$ \cite{tramer2018ensemble}. We craft AE on the ResNet50 surrogate model with $\epsilon=16/255$, and use DIM+TIM as the ``backbone'' to apply GNP. The results are presented in \tref{tab:3}, where we can see that GNP again boosts ASR significantly against the six ``secured'' models, achieving consistent performance improvements of 11.46--14.37\%.

\begin{figure}[ht!]
    \centering
    \includegraphics[width=.95\linewidth]{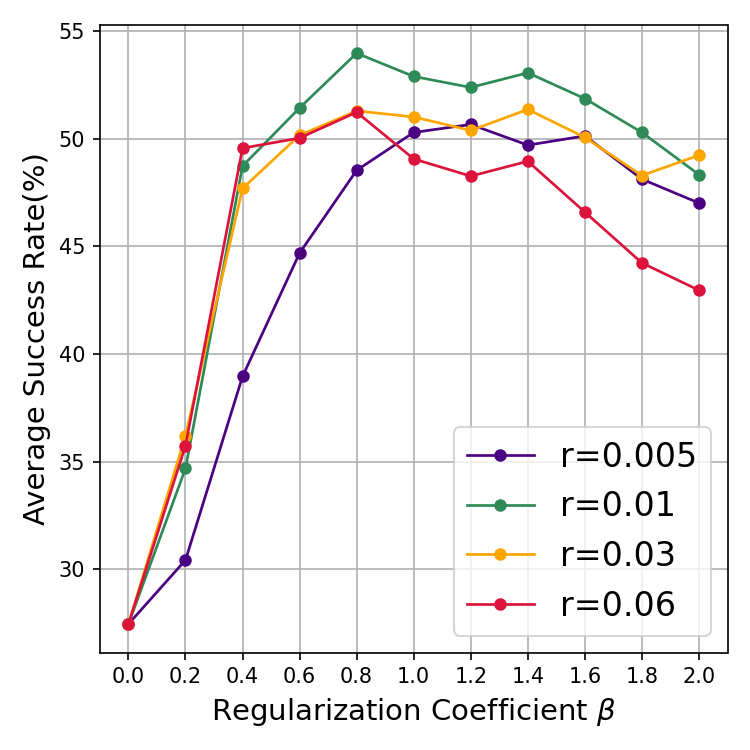}\vspace{-4mm}
    \caption{Avg. ASR under different hyperparameters $r$ and $\beta$.}
    \label{fig:ablation}
    \vspace{-5mm}
\end{figure}

\vspace{-2mm}
\subsection{Ablation Study}
\vspace{-2mm}
We conduct ablation study on the hyper-parameters of the proposed GNP attack, i.e., step length $r$ and regularization coefficient $\beta$. Since $r$ represents the radius of neighborhood that is flat around current AE, a larger $r$ is preferred; on the other hand, setting it too large will increase the approximation error of Taylor expansion and thus mislead the AE update direction. The $\beta$ is to balance the goal of fooling the surrogate model and finding flat optima. \fref{fig:ablation} reports the results of our ablation study, where ASR is averaged over 10 target models (excluding the source ResNet50) attacked by I-FGSM + GNP with $\epsilon=8/255$. We observe that adding the GNP regularization term clearly improves performance (as compared to $\beta=0$) and the performance gain is rather consistent for $\beta$ in a wide range of 0.6--1.6. The step length $r$ does not affect the performance gain too much either, and $r=0.01$ seems to be the most stable. Thus, the ablation study reveals that GNP is not hyper-parameter sensitive and works well in a variety of conditions.

\vspace{-2mm}
\section{Conclusion}  \label{sec:conclusion} 
\vspace{-2mm}

In this paper, we have proposed a new method for improving the transferability of AE from an optimization perspective, by seeking AE located at flat optima. We achieve this by introducing an input gradient norm penalty (GNP) which guides the AE search toward flat regions of the loss function. This GNP method is very flexible as it can be used with any gradient based AE generation methods. We conduct comprehensive experimental study and demonstrate that our method can boost the transferability of AE significantly.

This paper focuses on untargeted attacks, but GNP can be rather easily applied to targeted attacks as well, by making a small change to the loss function. We plan to have a thorough investigation in future work.


\bibliographystyle{IEEEbib}
\bibliography{ref}

\end{document}